# GhazalBench: Canonical Verse Access in LLMs across Persian Ghazals and Shakespearean Sonnets

**Ghazal Kalhor**[1] **Yadollah Yaghoobzadeh**[1,2]
[1]School of Electrical and Computer Engineering,
College of Engineering, University of Tehran, Tehran, Iran
[2]Tehran Institute for Advanced Studies, Khatam University, Tehran, Iran
**Correspondence:** kalhor.ghazal@ut.ac.ir, y.yaghoobzadeh@ut.ac.ir

## Abstract

Persian poetry plays an active role in Iranian cultural practice, where verses by canonical poets such as Hafez and Saadi are frequently quoted, paraphrased, or completed from incomplete cues. Supporting such interactions requires language models to reliably access canonical verses from semantic and lexical information. We introduce GhazalBench, a benchmark for evaluating how large language models (LLMs) access the canonical surface forms of Persian ghazal under usage-grounded conditions. Unlike prior work that primarily treats memorization as a liability, GhazalBench studies settings in which access to exact wording is functionally important. The benchmark evaluates completion and recognition under varied semantic and lexical cues. Across proprietary and open-weight multilingual LLMs, exact completion remains challenging, while recognition is substantially stronger. Performance also varies across poets and model families. Parallel experiments on Shakespearean sonnets yield markedly higher completion for several models, consistent with differences in exposure to canonical texts. Additional experiments suggest that post-training may reduce direct continuation-based access to canonical text. Our findings motivate evaluation frameworks that distinguish production from recognition and assess access to culturally significant canonical texts. GhazalBench is available at https://github.com/kalhorghazal/GhazalBench.

## 1 Introduction

Large language models (LLMs) are increasingly deployed in multilingual and culturally diverse settings, where successful interaction requires more than general linguistic fluency (Kim et al., 2025). Models are also expected to engage with *culturally grounded language practices* rooted in shared history, tradition, and collective memory (Hershcovich et al., 2022; Joshi et al., 2020). Although prior work has largely framed verbatim reproduction as a risk, emphasizing privacy leakage and unintended reproduction of training data (Carlini et al., 2021, 2022), some domains require reliable access to fixed canonical expressions. Interaction with culturally significant texts may depend on recognizing, completing, or reproducing them from partial or indirect cues. Whether LLMs can support such interactions remains underexplored.

Classical Persian poetry provides a salient example. Poets such as Hafez and Saadi remain enduring cultural references in Persian-speaking societies, particularly in Iran (Foltz, 2016; Shiran, 2020). Their verses are routinely quoted, partially recited, paraphrased, and alluded to in conversation and social rituals (Ordoubadian et al., 2006; Thackston et al., 2008). Such interactions are often initiated by partial semantic or lexical cues, requiring speakers to recognize or complete the canonical wording from fragmentary information.

Existing evaluations of LLMs on poetry have mainly emphasized creative generation, stylistic imitation, and aesthetic quality (Weatherby et al., 2024; Walsh et al., 2024b; Heuser, 2025). These settings do not capture the role of poetry as a shared cultural resource. The complementary question of how models access canonical poetic forms under usage-grounded conditions therefore remains largely unexplored.

We introduce GhazalBench, a benchmark for evaluating access to canonical verses from the ghazals of Hafez and Saadi. Rather than asking whether models can generate poetry, we examine whether they can complete or identify a canonical second verse from the poet's name, the first verse, and semantic or lexical cues. We intentionally study poems likely encountered during pretraining to examine how LLMs encode and access culturally significant literary knowledge through their parameters. We evaluate open-ended completion and multiple-choice recognition, drawing inspiration from the distinction between recall and recognition in cognitive psychology (Tulving et al., 1972; Tulving and Thomson, 1973). This design reveals how task formulation and cue availability shape access to canonical poetic form.

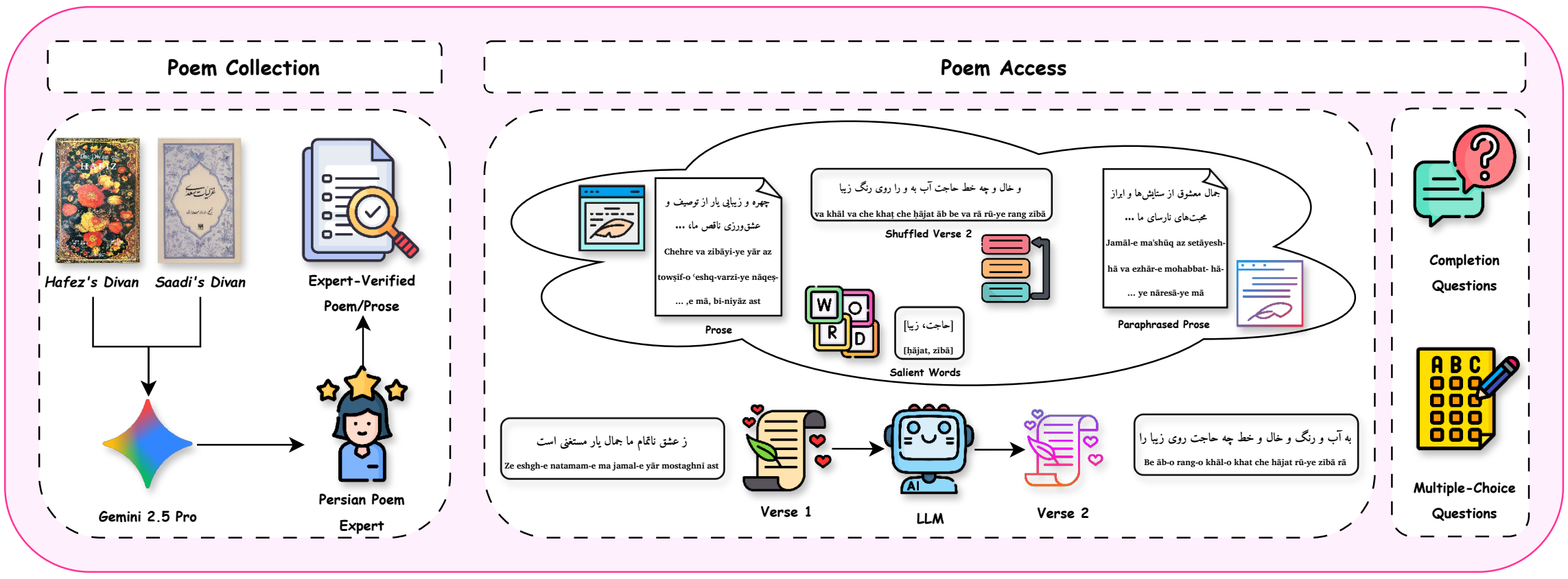

Figure 1: Pipeline illustrating the steps for poem collection and assessment of second-verse access (completion or recognition) given the first verse and different cues.

Experiments with proprietary and open-weight multilingual LLMs reveal a pronounced completion–recognition gap. Most models struggle to generate open-ended continuations, yet perform substantially better when the target verse is presented among alternatives. Performance also varies considerably across poets and model families. Parallel experiments on Shakespearean sonnets yield substantially higher completion rates for several models, consistent with differences in exposure to canonical texts, although language, poetic form, and post-training may also contribute.

This paper makes the following contributions: (i) We introduce GhazalBench, a usage-grounded benchmark for evaluating cue-dependent access to canonical Persian ghazals by Hafez and Saadi. (ii) We design complementary completion and recognition tasks that systematically vary semantic, sparse lexical, and word-order cues. (iii) We reveal a substantial completion–recognition gap across proprietary and open-weight multilingual LLMs, showing that canonical continuations are often recognizable but not readily generated. (iv) We demonstrate substantial variation across poets, model families, and languages, highlighting the role of corpus- and model-specific factors in canonical surface-form access.

## 2 Related Work

**Memorization and surface-form access in LLMs.** Memorization in LLMs has been studied mainly in relation to privacy leakage, unintended reproduction of training data, and overfitting (Carlini et al., 2021, 2022). Retrieval-augmented generation (RAG) offers an alternative by providing external documents at inference time (Lewis et al., 2020; Izacard et al., 2022), but this work largely concerns factual knowledge and document retrieval.

Recent studies have also examined culturally canonical texts. Zhang et al. (2024) analyze verbatim memorization of English and French literature; Attia et al. (2026) and Mensa et al. (2025) study proverb completion in Arabic and Italian; and poetry-focused work evaluates completion from partial or overlapping cues (D'Souza and Mimno, 2023; Chen et al., 2025; Walsh et al., 2024a). These studies primarily ask whether models retain canonical texts. In contrast, we study whether such texts are accessible under usage-grounded interaction, systematically varying semantic and lexical cues to the target verse.

**Poetry understanding and evaluation.** Prior work has evaluated LLMs' understanding and explanation of poetry through automatic metrics, LLM-as-Judge frameworks, and human assessment (Ma et al., 2025). Existing benchmarks cover poetic form in English sonnets (Walsh et al., 2024a) and semantic understanding in multilingual settings, including Arabic poetry (Al Ghallabi et al., 2025). They do not, however, systematically examine cue-dependent access to canonical poetic form.

**Poetry generation.** A substantial literature studies poetry generation with LLMs, focusing on style, rhyme, meter, and creativity across languages (Walsh et al., 2024b; Chen et al., 2024; Hernández Martínez, 2025). Our work instead evaluates access to existing canonical poetry rather than the generation of new poems.

**Existing persian benchmarks.** Recent Persian benchmarks evaluate language understanding, cultural knowledge, and literary content, including literary comprehension (Ghahroodi et al., 2024), allusion understanding (Nobakhtian et al., 2025), literary creativity (Tourajmehr et al., 2025), and proverb understanding and generation (Hosseini et al., 2026; Kalhor and Bahrak, 2026). GhazalBench complements these efforts by evaluating verse-level access to canonical poetry under usage-grounded cueing conditions.

## 3 Benchmark Construction

We first describe the poem collection process, followed by the task design and the cueing conditions used in the benchmark.

### 3.1 Poem Collection

To construct GhazalBench, we collect ghazals by Hafez and Saadi, two of the most widely recognized poets in the Persian literary tradition. Their verses remain deeply embedded in contemporary Iranian cultural practice and are frequently quoted, paraphrased, and completed in everyday interaction. Their ghazals therefore provide a natural testbed for studying LLM access to culturally canonical poetic forms. A ghazal is a classical poetic form composed of thematically related but structurally independent couplets, each consisting of two verses that follow a fixed meter and rhyme scheme (de Bruijn, 2000). Appendix A provides a brief overview of this form.

For Hafez, we extract the first 100 ghazals from the *Divan of Hafez* (Shirazi, 2009). The ordering of ghazals in the *Divan* is not based on popularity, chronology, or literary significance; rather, the poems are arranged alphabetically according to the final letter of their rhyme.

Each Hafez ghazal is collected couplet by couplet together with its corresponding prose explanation. We use Gemini 2.5 Pro (Comanici et al., 2025), a model with strong Persian-language capabilities,[1] to extract the ghazal number, couplet number, couplet text, and the reference prose explanations provided in Shirazi (2009).

For Saadi, we similarly collect the first 100 ghazals from his *Divan* (Shirazi, 2000), which is also arranged alphabetically according to the final letter of each ghazal's rhyme. We follow the same extraction procedure, using Gemini 2.5 Pro to obtain the ghazal number, couplet number, couplet text, and prose explanation from Barzegar Khaleghi and Oghdaei (2022).

All extracted outputs are manually verified against the original printed editions by authors proficient in Persian poetry, and any transcription or alignment errors are corrected. We additionally obtain the poetic meter of each ghazal from Ganjoor, a widely used digital repository of Persian poetry.[2]

For the experiments reported in this paper, we exclude ghazals for which no other ghazal by the same poet shares the same meter, as such matches are needed to construct metrically controlled distractors for recognition. Due to evaluation cost, we randomly sampled 50 ghazals from each corpus, yielding 438 Hafez couplets and 498 Saadi couplets. Since evaluation is conducted at the couplet level, these samples provide sufficient coverage of the main performance patterns.[3] Detailed statistics are reported in Appendix B, and Figure 1 summarizes the benchmark construction and evaluation pipeline.

### 3.2 Task Design

Each benchmark instance targets the second verse of a couplet. Across all conditions, the model receives the **poet's name** and the **first verse** of the target couplet. This reflects a common quotation-based interaction in which a speaker begins a familiar couplet and expects the listener to continue or recognize it. We then add semantic or lexical cues and examine how they affect access to the canonical continuation.

We evaluate two complementary task formulations. In **completion**, the model directly generates the canonical second verse. In **recognition**, it selects the correct verse from a set of candidates. Completion provides a more demanding test of open-ended access, whereas recognition tests whether the continuation becomes accessible when alternatives are provided.

[1] https://mcinext-mizan-llm-leaderboard.hf.space/

[2] https://ganjoor.net/

[3] As a stability check, we repeated the Saadi evaluation using all 100 collected ghazals instead of the sampled 50; the resulting scores changed only marginally.

| Condition | Additional Cue | Completion | Recognition |
|---|---|---|---|
| Baseline | None | ✓ | ✓ |
| Prose-mediated | Original prose explanation | ✓ | ✓ |
| Prose-paraphrase | Paraphrased prose explanation | ✓ | ✓ |
| Fragment-conditioned | Two salient target words | ✓ | – |
| Shuffled-form | Shuffled target verse | ✓ | ✓ |

Table 1: Cueing conditions in GhazalBench. All conditions additionally include the poet's name and the first verse of the target couplet.

Each recognition instance contains the correct verse and two distractors: one from another couplet in the *same ghazal*, and one from a *different ghazal by the same poet*. Because all couplets within a ghazal share the same meter, the same-ghazal distractor is automatically meter-matched. The different-ghazal distractor is selected from ghazals sharing the target meter.

Among metrically compatible candidates, we choose the verse with the highest semantic similarity to the correct continuation, measured by cosine similarity between embeddings from *paraphrase-multilingual-mpnet-base-v2* (Reimers and Gurevych, 2019). Controlling for meter and semantic proximity reduces the likelihood that models solve the task through coarse formal or semantic elimination. Answer options are randomly shuffled to mitigate positional bias (Pezeshkpour and Hruschka, 2024).

Table 1 summarizes the benchmark conditions. The no-cue condition serves as the baseline, while the remaining conditions add semantic or lexical information.

### 3.3 Cueing Conditions

The cueing conditions are designed to vary the amount and type of information available about the target verse while keeping the poet, first verse, and canonical continuation fixed. They are diagnostic rather than direct simulations of human memory: their purpose is to distinguish the contributions of semantic information, sparse lexical identity, and word order to canonical surface-form access.

**Baseline.** The baseline provides only the poet's name and the first verse. In completion, the model generates the second verse; in recognition, it selects the correct continuation from the metrically and semantically controlled candidates. All other conditions are compared with this baseline.

**Prose-mediated access.** The base input is augmented with a prose explanation of the target couplet, providing its meaning without the exact poetic wording. Comparison with the baseline tests whether semantic information facilitates completion or recognition of the canonical continuation.

**Prose-paraphrase robustness.** To distinguish semantic mediation from surface-level lexical overlap, we replace the original prose explanation with a paraphrased version. Gemini 3 Flash (Google LLC, 2025) is instructed to preserve the meaning of each explanation while substantially changing its vocabulary, phrasing, and sentence structure.

We assess paraphrase quality using BERTScore F1 for semantic similarity and Self-BLEU (Post, 2018) for lexical overlap with the original explanations. BERTScore is computed at the sentence level and averaged across the dataset, while Self-BLEU is computed at the corpus level. The resulting BERTScore F1 of 0.906 indicates high semantic fidelity, whereas the Self-BLEU score of 2.026 indicates limited surface overlap. Comparing the original and paraphrased prose conditions therefore helps determine whether performance gains arise primarily from preserved meaning or from shared lexical material.

**Fragment-conditioned access.** The base input is augmented with two salient words from the target second verse. These words provide sparse lexical evidence while withholding the verse's full syntax, word order, and most of its surface form. Gemini 3 Flash is instructed to select words that are maximally informative and memorable, approximating fragments that a speaker might remember when the remainder of the verse is unavailable.

This condition is evaluated only through completion. Providing target words in a recognition setting would make candidate selection substantially easier and would reduce the diagnostic value of the condition.

**Shuffled-form access.** The base input is augmented with a word-shuffled version of the target second verse. This condition preserves the com-plete lexical content of the target while disrupting its canonical order.

In completion, the model reconstructs the verse in its canonical order. In recognition, the task is reduced to a binary choice between the canonical verse and a shuffled version containing exactly the same words, therefore controlling for lexical identity and broad semantic content while testing sensitivity to canonical sequencing.

Success should not be interpreted solely as access to a stored canonical representation. Since the full lexical content is provided, models may also rely on grammar, poetic meter, and syntactic regularities. The condition therefore reflects both canonical-form access and broader language-modeling ability.

## 4 Experimental Setup and Results

**Research questions.** Our experiments address the following research questions: (i) RQ1: Model comparison. How does canonical surface-form access vary across commercial and open-weight LLMs from different model families and scales? (ii) RQ2: Task formulation. How does performance differ between open-ended completion and multiple-choice recognition? (iii) RQ3: Cue availability. How do semantic and sparse lexical cues affect access to canonical continuations? (iv) RQ4: Canonical ordering. How well can models reconstruct and recognize canonical word order when the complete lexical content of the target verse is provided? (v) RQ5: Post-training. How does canonical surface-form access differ between selected model variants with different post-training procedures? (vi) RQ6: Generalization. To what extent do the observed patterns generalize across Persian poets and to canonical poetry in English?

### 4.1 Models

We evaluate a diverse set of commercial and open-weight multilingual LLMs. The commercial models include GPT-5.2 (OpenAI, 2025) and Claude Sonnet 4.5 (Anthropic, 2025). The open-weight instruction-tuned models include DeepSeek V3.2 (Liu et al., 2025), DeepSeek R1 (DeepSeek-AI, 2025), Gemma 3 12B Instruct, Gemma 3 27B Instruct (Gemma Team and Google DeepMind, 2025), and Llama 4 Scout (AI, 2025).

These models were selected to cover commercial and open-weight systems, different model families and scales, and both conventional instruction-tuned and reasoning-oriented models. This diversity allows us to examine whether canonical surface-form access is associated with model scale, model family, or post-training strategy.

### 4.2 Inference and Prompting

Prompt templates for all cueing conditions and task formulations are provided in Appendix C. For all prompts used in both benchmark construction and model evaluation, we employ zero-shot prompting and run each prompt independently, without enabling web search, to ensure that the models rely solely on their internal knowledge. Moreover, as the focus of our study is accuracy rather than creativity, we set the temperature to 0 and the top-$p$ value to 1. Examples of each prompt used in the study are provided in Appendix C.

### 4.3 Main Results

#### 4.3.1 Completion

**Evaluation.** We first apply Persian-specific normalization to both model predictions and reference verses. This includes converting Arabic character variants to their Persian equivalents (e.g., 'ي' → 'ی' and 'ك' → 'ک'), removing diacritics, zero-width non-joiners, and kashida characters, and normalizing whitespace.

After normalization, a prediction is classified as an *exact match* if its Levenshtein edit distance from the reference is at most one, two, or three characters for verses shorter than 40 characters, between 40 and 79 characters, or at least 80 characters, respectively. This length-dependent criterion accommodates minor typographical variation while still requiring near-verbatim reproduction of the canonical verse.

For predictions that do not satisfy the exact-match criterion, we compute the Character Error Rate (CER), defined as the Levenshtein distance normalized by the reference length. Predictions with CER $\leq$ 0.2 are classified as *partial matches*, while all remaining predictions are labeled *no match*. We selected the threshold of 0.2 by comparing candidate thresholds of 0.1, 0.15, 0.2, and 0.25 against human judgments of near-correct completion.

**Results.** Table 2, "Completion" section, reports the percentage of successful completions, combining exact and partial matches. Representative outputs and common failure modes are provided in Appendix D. The most common failure modes include hallucinated continuations, prosaic paraphrases, incomplete completions, and contextual confusion, where a verse from another ghazal is produced. The results show that open-ended access to canonical continuations remains difficult for most models. Under the baseline condition, GPT-5.2 achieves substantially higher completion performance than all other models on Hafez, reaching 44.5%. DeepSeek R1 and DeepSeek V3.2 form a distant second group, while Claude Sonnet 4.5, Llama 4, and both Gemma 3 models achieve only limited success.

Adding the original prose explanation improves Hafez completion for GPT-5.2 and Claude Sonnet 4.5, with smaller gains for the DeepSeek models. These gains diminish after paraphrasing. For example, GPT-5.2 rises from 44.5% at baseline to 51.6% with the original prose, but returns to 44.5% with paraphrased prose. This suggests that prose cues aid access partly through lexical correspondence rather than meaning alone.

Providing two salient target words is less effective, yielding little improvement for most models and substantially reducing GPT-5.2 performance. Sparse lexical fragments therefore appear insufficient to reliably activate the canonical continuation.

The shuffled-form condition shows a different pattern. It yields the highest completion performance for Claude Sonnet 4.5, DeepSeek V3.2, and DeepSeek R1, but not for GPT-5.2 or the Gemma models. The large gains for the DeepSeek models suggest that preserving the full lexical inventory substantially aids reconstruction, even when canonical order is disrupted. However, because all target words are provided, success may reflect a combination of canonical access, grammatical reconstruction, and sensitivity to poetic structure.

DeepSeek R1 performs slightly better than DeepSeek V3.2 across most Hafez conditions, with the clearest advantages under salient-word and shuffled-form cueing. This suggests that the two post-training regimes differ in how effectively they exploit sparse or structurally disrupted cues, although the present results do not establish the mechanism responsible for this difference.

Performance also varies sharply across poets.[4] On Saadi, GPT-5.2 achieves only 2.6% at baseline, compared with 44.5% on Hafez, while prose cueing raises it to 10.8%. The DeepSeek models also remain near zero under baseline, prose, paraphrase, and salient-word conditions, but improve with shuffled-form cueing. This large within-language gap shows that language-level resource availability alone cannot explain canonical access; exposure may vary substantially across literary corpora within the same language and tradition. Since the models' training data and post-training pipelines are unavailable, however, the source of these differences remains uncertain.

Overall, except for GPT-5.2 on Hafez, open-ended completion remains weak even when additional cues are supplied. The results also show that cue effectiveness is model-dependent: prose is most helpful for GPT-5.2 and Claude, whereas shuffled lexical content provides the greatest benefit to the DeepSeek models.

#### 4.3.2 Recognition

Table 2, under "Recognition," reports recognition accuracy. All models perform substantially better when the canonical continuation is presented among candidates than when it must be generated directly. This gap is especially pronounced for Gemma 3 and Llama 4, which show near-zero completion but well-above-chance recognition.

Under the baseline condition, GPT-5.2 performs best, reaching 92.9% on Hafez and 85.5% on Saadi. Other models achieve more moderate baseline accuracy, ranging from 46.3% to 66.0% on Hafez. Thus, even models unable to reproduce the second verse may retain enough information to distinguish it from metrically and semantically controlled alternatives. Prose explanations substantially improve recognition for every evaluated model. On Hafez, most models exceed 90% accuracy with the original explanation, except Llama 4 at 74.9%. Paraphrased explanations yield similarly strong results, though usually slightly lower. The small decline after paraphrasing suggests that semantic information drives much of the gain, while lexical overlap provides an additional benefit.

Error patterns further show that recognition is not based solely on meter. When incorrect, models disproportionately select the distractor from the same ghazal rather than the metrically matched distractor from another ghazal (Figures 2 and 3 in Appendix). Since both are meter-controlled, this preference suggests sensitivity to thematic, lexical, or stylistic coherence.

[4] Due to API cost constraints, the Saadi evaluation covers three representative models: GPT-5.2, DeepSeek V3.2, and DeepSeek R1, selected as the strongest commercial model and two open-weight models with relatively high Hafez completion.

| Model | Completion | | | | | Recognition | | | |
|---|---|---|---|---|---|---|---|---|---|
| | Baseline | +Prose | +Para. | +Salient | +Shuffle† | Baseline | +Prose | +Para. | +Shuffle† |
| **Hafez** | | | | | | | | | |
| GPT-5.2 | **44.5** | **51.6** | **44.5** | **33.8** | **23.1** | **92.9** | **100.0** | **98.9** | **99.5** |
| Sonnet 4.5 | 3.6 | 7.1 | 4.6 | 4.6 | 11.9 | 66.0 | 99.3 | 96.1 | 95.4 |
| DeepSeek V3.2 | 7.1 | 7.3 | 5.3 | 3.2 | 11.7 | 50.9 | 92.5 | 88.6 | 79.0 |
| DeepSeek R1 | 9.6 | 10.5 | 7.3 | 8.2 | 13.0 | 58.9 | 97.3 | 91.8 | 92.0 |
| Gemma 3 27B | 0.2 | 0.2 | 0.2 | 0.2 | 1.4 | 51.6 | 95.2 | 90.0 | 86.8 |
| Gemma 3 12B | 0.2 | 0.2 | 0.0 | 0.0 | 0.2 | 49.1 | 90.0 | 89.5 | 65.3 |
| Llama 4 | 0.7 | 1.6 | 1.4 | 0.7 | 1.6 | 46.3 | 74.9 | 80.8 | 77.4 |
| **Saadi** | | | | | | | | | |
| GPT-5.2 | **2.6** | **10.8** | **6.6** | **7.6** | **15.9** | **85.5** | **99.8** | **98.4** | **98.4** |
| DeepSeek V3.2 | 0.4 | 0.6 | 0.0 | 0.6 | 5.8 | 56.0 | 93.6 | 91.0 | 62.9 |
| DeepSeek R1 | 0.4 | 1.4 | 0.6 | 0.4 | 8.4 | 61.2 | 97.0 | 95.8 | 90.4 |

Table 2: Completion and recognition performance (%) on Hafez and Saadi. Completion scores combine exact and partial matches. The baseline provides the poet's name and first verse; the prose, paraphrase, and salient-word conditions add partial semantic or lexical cues. Recognition under the baseline and prose-based conditions uses three-way multiple choice. The shuffled-form condition (†) provides the complete target vocabulary in disrupted order; its recognition variant is binary and is therefore not directly comparable with the other recognition columns. Saadi results are reported for three representative models because of API cost constraints.

The shuffled-form condition tests a different capability by presenting a binary choice between the canonical verse and the same words in disrupted order. All models perform above chance, indicating sensitivity to canonical sequencing beyond lexical identity. GPT-5.2 performs near ceiling on both poets, while Claude Sonnet 4.5 and DeepSeek R1 also achieve strong results on Hafez.

DeepSeek R1 consistently outperforms DeepSeek V3.2 in recognition. The gap is largest in the shuffled-form condition, where R1 reaches 92.0% (90.4%) compared with 79.0% (62.9%) for V3.2 on Hafez (Saadi). This suggests that R1 more effectively exploits structural information when lexical content is fixed, although further work is needed to determine whether the advantage stems from reasoning-oriented post-training.

The Hafez–Saadi comparison also reveals substantial poet-level variation. GPT-5.2 remains strong on both poets, though its baseline recognition declines from 92.9% on Hafez to 85.5% on Saadi. DeepSeek V3.2 performs slightly better on the Saadi baseline but markedly worse in the shuffled-form condition. These differences suggest that access to and representation of canonical verse vary across poetic corpora rather than reflecting a uniform Persian-language capability.

Overall, Table 2 reveals a robust completion–recognition gap. Many models identify canonical continuations accurately, especially with semantic cues, yet fail to generate them directly. Canonical knowledge may therefore be accessible in recognition without being readily elicited through open-ended completion.

**Statistical comparison of DeepSeek variants.** Because DeepSeek V3.2 and DeepSeek R1 offer the closest comparison of post-training regimes and are evaluated on the same instances, we apply pairwise McNemar tests (McNemar, 1947). Tables 24 and 25 in Appendix F report the results. Across the full set of models, most pairwise differences are statistically significant, although smaller differences are not consistently significant. For completion, significant differences between the DeepSeek variants appear only in some Hafez conditions and none for Saadi. Recognition shows clearer differences, significant in most Hafez conditions and all Saadi conditions. We restrict our interpretation to the DeepSeek pair because the remaining models differ more substantially in architecture, scale, training data, and coverage.

### 4.4 Cross-Lingual Analysis: Shakespeare's Verse Completion

To examine whether weak Persian completion reflects a general limitation of current LLMs or language- and corpus-specific factors, we conduct parallel experiments on *Shakespeare's Sonnets* (Shakespeare, 2002). The sonnets form a widely circulated canonical corpus with stable text and strong formal constraints, providing a useful high-resource comparison (Gray et al., 1995). Although sonnets and ghazals differ structurally and historically, both support interactions involving exact poetic continuation.

We randomly sample 50 of the 154 sonnets, yielding 350 consecutive line pairs. Models receive Shakespeare's name and the first line and must generate the following line. We evaluate the baseline, salient-word, and shuffled-form conditions; prose conditions are omitted because comparable reference explanations are unavailable. Salient words are selected using Gemini 3 Flash, and sentence-initial capitalization is removed from salient and shuffled cues to prevent trivial positional clues.

Appendix E compares basic properties of the English and Persian datasets. They have similar word counts and lexical diversity, although they differ in character length, vocabulary size, poetic form, and other linguistic properties. These differences could be partly explained by the larger size of the Persian corpus (Hafez and Saadi combined). The analysis should therefore be interpreted as diagnostic rather than as a fully controlled cross-lingual comparison.

As shown in Table 3, most models complete Shakespearean lines substantially more accurately than Persian verses. GPT-5.2, Claude Sonnet 4.5, and both DeepSeek models achieve high English baseline accuracy, suggesting that the low Persian scores cannot be explained solely by the difficulty of exact poetic continuation. Greater exposure to canonical English poetry is a plausible explanation, although inaccessible pretraining data prevents direct confirmation.

The pattern is not uniform: Gemma 3 and Llama 4 remain weak in English, indicating substantial differences across model families. Shuffled-form cues generally produce the highest scores, whereas salient words yield smaller and less consistent gains. These results again show that full lexical identity is considerably more useful than a small number of lexical fragments.

| Model | Baseline | +Salient | +Shuffle |
|---|---|---|---|
| GPT-5.2 | 92.6 | **97.7** | 98.0 |
| Sonnet 4.5 | **96.3** | 94.0 | **98.9** |
| DeepSeek V3.2 | 80.3 | 60.0 | 81.1 |
| DeepSeek R1 | 91.7 | 94.3 | 97.1 |
| Gemma 3 27B | 6.0 | 5.4 | 20.6 |
| Gemma 3 12B | 0.6 | 1.2 | 6.0 |
| Llama 4 | 19.7 | 20.3 | 24.0 |

Table 3: Successful second line completion percentages in Shakespeare's sonnets.

**Instruction tuning effect.** Our main results suggest that post-training may influence canonical surface-form access, as DeepSeek R1 and DeepSeek V3.2 show different patterns across completion and recognition conditions. Prior work also indicates that instruction tuning can weaken some pretraining-derived behaviors, including transfer to unseen tasks and few-shot in-context learning (Mueller et al., 2024; Irsoy et al., 2025). To examine this possibility more directly, we conduct an additional experiment on Shakespearean sonnets. Because matched base checkpoints are unavailable for the model series in our main evaluation, we use two pairs of base and instruction-tuned models: Qwen2.5-7B and Qwen2.5-7B-Instruct (Qwen et al., 2025), and Gemma2-9B and Gemma2-9B-Instruct (Team et al., 2024).

We use model-appropriate prompts: the instruction-tuned models receive our standard completion instruction, whereas the base models receives only Shakespeare's name and the first line, with the following generated line taken as its prediction. For Qwen2.5-7B, the base model achieves an 8.3% completion rate, compared with 2.0% for the instruction-tuned model. Similarly, the base Gemma2-9B model achieves a completion rate of 20.6%, compared with 4.6% for its instruction-tuned counterpart. These results suggest that post-training may reduce direct continuation-based access, possibly by shifting behavior from next-token continuation toward instruction-mediated generation. However, because prompt format also differs and the analysis covers only two model pairs, the result should be considered preliminary rather than causal evidence.

## 5 Discussion and Conclusion

We introduced GhazalBench, a benchmark for evaluating cue-dependent access to culturally canonical Persian poetry. It tests whether LLMs can complete, recognize, and reconstruct canonical verses from Hafez and Saadi under controlled semantic and lexical cueing conditions.

Our main finding is a substantial completion–recognition gap. Most models struggle to generate canonical continuations but identify them accurately when presented among alternatives, particularly when semantic cues are available. Cue effectiveness also varies by model family: prose explanations benefit some models, while complete shuffled lexical content is more useful for others.

Sparse salient words are generally insufficient.

Performance also varies considerably across poets and languages. Completion is markedly weaker for Saadi than for Hafez, while several models perform substantially better on Shakespearean sonnets than on Persian ghazals. These differences may reflect unequal exposure to individual canonical corpora, while language, poetic form, post-training, and model design may also contribute; additional experiments with matched base and instruction-tuned models provide preliminary evidence that post-training can affect direct continuation-based access.

More broadly, strong recognition does not imply reliable open-ended access to exact canonical text. Recognition-only evaluations may therefore overestimate models' ability to participate in quotation-based and allusive cultural interactions. GhazalBench provides a controlled framework for studying this distinction across models, cue types, poets, and languages.

A curated retrieval-augmented system (RAG) could likely resolve many instances by searching an external poetry collection. GhazalBench does not treat parametric access as a replacement for retrieval; rather, it characterizes the capabilities and limitations of standalone LLMs and helps identify when external resources are needed. Future work could extend the benchmark to RAG settings.

## Limitations

GhazalBench is limited in scope to ghazals by Hafez and Saadi and evaluates only second-verse access. Although these poets are highly canonical, the results may not generalize to other Persian poets, poetic forms, or culturally significant texts. We additionally evaluate only a random sample of 50 ghazals per poet, and the Saadi experiments are restricted to three representative models because of API cost constraints. Future work should extend the benchmark to additional poets, larger collections, and other literary traditions.

Our cross-lingual analysis is restricted to Shakespearean sonnets and should be interpreted as diagnostic rather than as a controlled language-level comparison. Persian ghazals and English sonnets differ in language, poetic structure, editorial history, cultural circulation, and likely training exposure. Since the pretraining corpora of most evaluated models are unavailable, we cannot determine whether performance differences arise from corpus frequency, data contamination, model architecture, post-training, or other factors. The observed differences between Hafez and Saadi further indicate that access may vary substantially across individual canonical corpora even within the same language and poetic form.

The benchmark evaluates behavioral access rather than directly measuring whether a verse was memorized during pretraining. Successful completion may reflect verbatim storage, learned poetic regularities, or both, while recognition may sometimes be supported by semantic, stylistic, or metrical compatibility rather than exact canonical knowledge. This is particularly important for the shuffled-form condition, where the complete lexical content is provided and models may rely on syntax, meter, and language modeling ability to reconstruct a plausible ordering.

Our post-training analysis is also preliminary. The comparison between DeepSeek V3.2 and DeepSeek R1 does not isolate a single post-training component, while the Qwen base–instruct comparison covers only one model family and a small Shakespeare subset. Moreover, the base and instruction-tuned checkpoints receive model-appropriate but different prompt formats, so the observed difference cannot be attributed causally to instruction tuning alone. More controlled comparisons across multiple matched model families and prompt formats are needed.

We use a single zero-shot prompt template per condition, and alternative prompting, decoding, or few-shot demonstrations may elicit different levels of canonical access.

Finally, we evaluate standalone LLMs without external retrieval. A retrieval-augmented system equipped with a reliable and comprehensive poetry collection could likely solve many instances more accurately. GhazalBench therefore should not be interpreted as arguing that parametric access should replace retrieval. Rather, it characterizes when canonical forms are accessible from model parameters alone and may help identify cases in which external retrieval is required. Future work could evaluate hybrid systems that convert incomplete semantic or lexical cues into effective retrieval queries.

## A Ghazal Introduction

The ghazal is a poetic form typically centered on romantic themes, consisting of five to twelve couplets. The first verse of the opening couplet rhymes with all subsequent even-numbered verses. Etymologically, ghazal means “talking with the beloved, declaring one’s love, recounting youth, and the conversation of companionship and love.” Table 4 provides some examples of Hafez’s ghazal couplets along with their English translations.

## B Benchmark Statistics

**Dataset overview.** GhazalBench is constructed at the couplet level and consists of 936 Persian couplets (438 by Hafez and 498 by Saadi) and 350 English couplets. Each couplet is evaluated across multiple task settings, resulting in a total of 9,474 prompts per model across all tasks. This evaluation design enables a systematic analysis of model behavior under different access conditions.

**Task distribution.** Table 5 summarizes the distribution of tasks across languages and evaluation settings.

**Couplet length.** We report the length of couplets measured in characters.

- **Persian:** mean = 64.2, std = 8.2
- **English:** mean = 85.7, std = 6.2

**Prompt length.** We further report the length of input prompts (in characters).

- **Persian (Completion):** mean = 936.4, std = 283.6
- **Persian (Recognition):** mean = 721.6, std = 113.9
- **English (Completion):** mean = 1250.2, std = 416.8

**Discussion.** While the dataset is moderate in size, the benchmark is designed to systematically vary access conditions at the couplet level. This results in a large number of evaluation instances per model and enables controlled comparisons across multiple task settings, particularly for analyzing completion, recognition, and semantic mediation in poetic access.

## C Example Prompts

In this section, we provide an example of each prompt used in this study. For the Persian prompts, we provide examples only for Hafez, as the template is similar for Saadi, with only the poet’s name differing. Additionally, each Persian prompt is accompanied by its English translation.

### C.1 Poem Collection Prompt

An example prompt used to collect ghazals and their respective prose, couplet by couplet, is provided in Table 6.

### C.2 Auxiliary Prompts

The following auxiliary prompts are used to construct canonical surface-form access questions and are not themselves used for model evaluation.

#### C.2.1 Prose Paraphrasing

#### C.2.2 Salient Words Selection

### C.3 Completion / Recognition Prompts

#### C.3.1 Completion Questions

Example prompts for completing Hafez’s ghazals at the verse level, given different input information, are provided in Tables 10–14.

#### C.3.2 Recognition Questions

Example prompts for verse-level recognition of Hafez’s ghazals under different input conditions are provided in Tables 15–18.

#### C.3.3 English Prompts

Example prompts for line completion of Shakespeare’s sonnets under different input conditions are provided in Tables 19–21.

| Couplet Text | English Translation |
|---|---|
| هرگز نمیرد آن که دلش زنده شد به عشق / ثبت است بر جریده عالم دوام ما | Never dies the one whose heart is quickened by love; Our everlasting life is inscribed in the world's record. |
| تا مرا عشق تو تعلیم سخن گفتن کرد / خلق را ورد زبان مدحت و تحسین من است | Ever since your love taught me the art of speech, My praise and admiration have become the constant refrain on people's tongues. |
| ندای عشق تو دیشب در اندرون دادند / فضایِ سینهٔ حافظ هنوز پر ز صداست | The call of your love was raised within my soul last night; The space of Hafez's chest is still filled with its echo. |
| راهیست راه عشق که هیچش کناره نیست / آنجا جز آن که جان بسپارند چاره نیست | The path of love is a road with no end in sight; There, there is no choice but to surrender one's soul. |
| لطیفه‌ایست نهانی که عشق از او خیزد / که نام آن نه لب لعل و خط زنگاریست | There is a hidden mystery from which love arises, A mystery not called ruby lips or the faint mustache of youth. |

Table 4: Examples of Hafez's ghazal couplets along with their English translations.

| Language | Task Type | # Instances | Proportion |
|---|---|---|---|
| Persian | Completion | 4,680 | 55.6% |
| Persian | Recognition | 3,744 | 44.4% |
| English | Completion | 1,050 | 100.0% |

Table 5: Task distribution across languages and evaluation settings.

شما یک مدل زبانی هستید و وظیفه دارید صرفاً متن کتاب «دیوان حافظ با شرح کامل ابیات» که در این فایل آپلود شده است را بخوانید و بدون هیچ‌گونه تغییر، حذف، بازنویسی یا افزودن، محتوای آن را استخراج نمایید.

لطفاً دستورالعمل‌های زیر را دقیقاً رعایت کنید:
1. هر غزل را به صورت بیت به بیت بررسی کنید.
2. برای هر بیت، شرح یا تفسیر مربوط به همان بیت را عیناً و بدون تغییر از متن کتاب استخراج و درج نمایید.
3. خروجی را در قالب یک جدول با چهار ستون زیر ارائه دهید:
- «شماره غزل»
- «شماره بیت»
- «متن بیت»
- «شرح یا تفسیر»
4. در ستون «متن بیت»، دو مصراع هر بیت را با علامت « / » از یکدیگر جدا کنید.
5. در ستون «شرح یا تفسیر» فقط و فقط متن کتاب را عیناً بیاورید و هیچ توضیح یا تحلیل شخصی خود را اضافه نکنید.
6. اگر برای بیتی شرحی در کتاب وجود ندارد، ستون مربوط به شرح را خالی بگذارید.
7. ساختار جدول را به گونه‌ای ارائه دهید که بتوان آن را به راحتی در فایل lecxE یا VSC ذخیره و پردازش نمود.

توجه: رعایت امانت‌داری کامل در نقل متن کتاب ضروری است.

**English Translation:**
You are a language model. Your task is to read only the text of the book *"Divan of Hafez with Complete Commentary on the Couplets"* provided in this uploaded file, and extract its content without any modification, omission, paraphrasing, or addition.

Please strictly follow these instructions:
1. Process each ghazal couplet by couplet.
2. For each couplet, extract and reproduce the corresponding prose exactly as it appears in the book, without any changes.
3. Present the output as a table with the following four columns:
- "Ghazal Number"
- "Couplet Number"
- "Couplet Text"
- "Prose"
4. In the "Couplet Text" column, separate the two verses of each couplet using the symbol " / ".
5. In the "Prose" column, include only the exact text from the book and do not add any personal explanation or interpretation.
6. If no prose is provided for a couplet in the book, leave the corresponding cell empty.
7. Structure the table so that it can be easily exported to and processed in Excel or CSV format.

Note: Strict fidelity to the original text is essential.

Table 6: Example prompt for collecting ghazals and their prose from the Divan of Hafez.

شما یک مدل زبانی هستید که برای بازنویسی نثر فارسی طراحی شده‌اید.

به شما یک متن فارسی داده می‌شود. وظیفه‌ی شما این است که آن متن را با حفظ کامل معنا بازنویسی کنید، به‌گونه‌ای که تا حد امکان واژگان، عبارات و ساختار جمله‌ها تغییر کنند، بدون آن‌که محتوا یا پیام متن دچار تغییر شود.

لطفاً دستورالعمل‌های زیر را دقیقاً رعایت کنید:
1- نثر بازنویسی‌شده باید از نظر معنایی کاملاً معادل نثر ورودی باشد.
2- نثر بازنویسی‌شده باید کاملاً طبیعی، روان و مطابق با فارسی معیار باشد.
3- هیچ‌گونه اطلاعات، توضیح یا محتوای جدیدی نباید به متن اضافه شود و هیچ بخشی از معنا نیز نباید حذف شود.
4- جایگزینی واژگان و عبارات باید به‌گونه‌ای انجام شود که شباهت واژگانی با متن ورودی حداقل شود، مشروط بر آن‌که معنا کاملاً حفظ شود.
5- ساختار جمله‌ها باید تا حد امکان تغییر کند، بدون تغییر در معنا.
6- نثر بازنویسی‌شده باید غیرشاعرانه و صرفاً توضیحی باقی بماند.
7- خروجی را فقط در قالب زیر بازگردانید و از ارائه هرگونه توضیح اضافی خودداری کنید:
<answer>نثر بازنویسی‌شده</answer>

نثر ورودی:
<prose>ای دل با این شتاب کجا می‌روی، فقط به سیب زنخدان یار نگاه مکن و فریفته‌اش مشو بلکه متوجه چاه ذقن او نیز باش که در راهت است.</prose>

**English Translation:**
You are a language model designed to paraphrase Persian prose.

You will be given a Persian text. Your task is to rewrite the text while fully preserving its meaning, such that the vocabulary, expressions, and sentence structure are changed as much as possible without altering the content or message.

Please strictly follow these instructions:
1- The rewritten prose must be semantically equivalent to the input.
2- The rewritten prose must be natural, fluent, and conform to standard Persian.
3- Do not add any new information, explanation, or content, and do not omit any part of the meaning.
4- Replace words and expressions so as to minimize lexical overlap with the input, while fully preserving the meaning.
5- Modify sentence structure as much as possible without changing the meaning.
6- The rewritten prose must remain non-poetic and purely explanatory.
7- Return the output strictly in the following format and do not include any additional text:
<answer>Paraphrased prose</answer>

Input prose:
<prose>O heart, where are you going in such haste? Do not fix your gaze only on the beloved's dimpled chin and be enchanted by it; also be mindful of the pit of the chin that lies in your path.</prose>

Table 7: Example prompt for paraphrasing prose from a Hafez ghazal at the couplet level.

شما یک مدل زبانی هستید که برای شناسایی واژه‌های برجسته از ابیات غزل‌های حافظ طراحی شده‌اید.

به شما مصراع دوم از یکی از ابیات غزل‌های حافظ داده می‌شود. وظیفه‌ی شما این است که دقیقاً دو واژه از همان مصراع را انتخاب کنید که از نظر معنایی و حافظه‌ای برجسته‌ترین باشند؛ به‌گونه‌ای که حتی در صورت فراموش شدن کل مصراع، بیشترین احتمال به خاطر سپرده شدن را داشته باشند.

لطفاً دستورالعمل‌های زیر را دقیقاً رعایت کنید:
1- واژه‌ها باید دقیقاً به همان صورت سطحی در مصراع ورودی وجود داشته باشند.
2- هیچ واژه‌ای را تغییر ندهید، خلاصه نکنید، یا بازنویسی نکنید.
3- فقط واژه‌های منفرد (نه عبارت یا ترکیب چندواژه‌ای) انتخاب کنید.
4- دو واژه‌ی انتخاب‌شده باید متمایز و غیرتکراری باشند، حتی اگر یک واژه بیش از یک‌بار در مصراع ظاهر شده باشد.
5- خروجی را فقط در قالب زیر بازگردانید و از ارائه هرگونه توضیح اضافی خودداری کنید:
["واژه اول", "واژه دوم"]

مصراع ورودی:
<second_verse>که عشق از پرده عصمت برون آرد زلیخا را</second_verse>

**English Translation:**
You are a language model designed to identify salient words from couplets of Hafez's ghazals.

You will be given the second verse of a couplet from a Hafez ghazal. Your task is to select exactly two words from that verse that are the most salient in terms of meaning and memorability, such that they are most likely to be remembered even if the entire verse is forgotten.

Please strictly follow these instructions:
1- The selected words must appear exactly in their surface form in the input verse.
2- Do not modify, shorten, or paraphrase any word.
3- Select only single words (not phrases or multi-word expressions).
4- The two selected words must be distinct and non-repeating, even if a word appears multiple times in the verse.
5- Return the output strictly in the following format and do not include any additional text:
["word1", "word2"]

Input verse:
<second_verse>Love would draw Zuleikha out from behind the veil of chastity</second_verse>

Table 8: Example prompt for choosing salient words from a Hafez ghazal at the verse level.

You are a language model designed to identify salient words from lines of Shakespeare's Sonnets.

You are given the second line of a couplet from one of Shakespeare's sonnets. Your task is to select exactly two words from that line that are the most semantically and memorably salient, such that even if the entire line were forgotten, these two words would be the most likely to be remembered.

Please strictly follow the instructions below:
1- The selected words must appear exactly in the input line in their original surface form.
2- Do not modify, paraphrase, lemmatize, or rewrite any word.
3- Select only single words (not phrases or multi-word expressions).
4- The two selected words must be distinct and non-redundant, even if a word appears more than once in the line.
5- Return the output strictly in the following format and do not provide any additional explanation:
["first_word", "second_word"]

Input line:
<second_line>When beauty lived and died as flowers do now,</second_line>

Table 9: Example prompt for choosing salient words from a Shakespeare sonnet at the line level.

شما یک مدل زبانی هستید که برای بازیابی مصراع دوم دقیق و اصیل حافظ از دیوان او طراحی شده‌اید.

به شما مصراع اول از یکی از ابیات غزل‌های حافظ داده می‌شود. وظیفه‌ی شما این است که مصراع دوم واقعی و موجود در دیوان حافظ را که مربوط به همین مصراع اول است، شناسایی کرده و بدون هیچ‌گونه تغییری بازگردانید.

لطفاً دستورالعمل‌های زیر را دقیقاً رعایت کنید:
1- هیچ مصراعی نسازید؛ فقط از متن معتبر دیوان حافظ بازیابی کنید.
2- مصراع باید کاملاً مطابق متن اصلی حافظ باشد؛ هیچ واژه یا نشانه‌ای را تغییر ندهید.
<answer>مصراع دوم واقعی حافظ</answer>

مصراع اول:
<first_verse>خوبان پارسی گو بخشندگان عمرند</first_verse>

**English Translation:**
You are a language model designed to retrieve the exact and authentic second verse of Hafez from his Divan.

You will be given the first verse of a couplet from a Hafez ghazal. Your task is to identify the true second verse from the Divan of Hafez that corresponds to this first verse, and return it without any modification.

Please strictly follow these instructions:
1- Do not generate any verse; only retrieve it from the authentic text of the Divan of Hafez.
2- The verse must exactly match the original text of Hafez; do not alter any word or punctuation.
<answer>Exact second verse</answer>

Input first verse:
<first_verse>The Persian-speaking beloveds are givers of life</first_verse>

Table 10: Example prompt for completing the second verse of a ghazal's couplet, given the poet's name (Hafez).

شما یک مدل زبانی هستید که برای بازیابی مصراع دوم دقیق و اصیل حافظ از دیوان او طراحی شده‌اید.

به شما مصراع اول یکی از ابیات غزل‌های حافظ داده می‌شود و سپس نثر روان فارسی همان بیت ارائه می‌گردد. وظیفه‌ی شما این است که مصراع دوم واقعی و موجود در دیوان حافظ را که مربوط به همین بیت است، شناسایی کرده و بدون هیچ‌گونه تغییری بازگردانید.

لطفاً دستورالعمل‌های زیر را دقیقاً رعایت کنید:
1- هیچ مصراعی نسازید؛ فقط از متن معتبر دیوان حافظ بازیابی کنید.
2- مصراع باید کاملاً مطابق متن اصلی حافظ باشد؛ هیچ واژه یا نشانه‌ای را تغییر ندهید.
<answer>مصراع دوم واقعی حافظ</answer>

مصراع اول:
<first_verse>غرور حسنت اجازت مگر نداد ای گل</first_verse>

نثر بیت:
<couplet_prose>ای گل، حتماً غرور زیبایی‌ات اجازه نداد که از حال بلبل عاشق خود احوالی بپرسی.</couplet_prose>

**English Translation:**
You are a language model designed to retrieve the exact and authentic second verse of Hafez from his Divan.

You will be given the first verse of a couplet from a Hafez ghazal, followed by a fluent Persian prose rendering of the same couplet. Your task is to identify the true second verse from the Divan of Hafez that corresponds to this couplet, and return it without any modification.

Please strictly follow these instructions:
1- Do not generate any verse; only retrieve it from the authentic text of the Divan of Hafez.
2- The verse must exactly match the original text of Hafez; do not alter any word or punctuation.
<answer>Exact second verse</answer>

Input first verse:
<first_verse>O flower, perhaps the pride of your beauty did not permit it</first_verse>

Couplet prose:
<couplet_prose>O flower, surely the pride in your beauty did not allow you to ask about the state of your lover nightingale.</couplet_prose>

Table 11: Example prompt for completing the second verse of a ghazal's couplet, given the poet's name (Hafez) and the couplet's prose.

شما یک مدل زبانی هستید که برای بازیابی مصراع دوم دقیق و اصیل حافظ از دیوان او طراحی شده‌اید.

به شما مصراع اول یکی از ابیات غزل‌های حافظ داده می‌شود و سپس نثر روان فارسی همان بیت ارائه می‌گردد. وظیفه‌ی شما این است که مصراع دوم واقعی و موجود در دیوان حافظ را که مربوط به همین بیت است، شناسایی کرده و بدون هیچ‌گونه تغییری بازگردانید.

لطفاً دستورالعمل‌های زیر را دقیقاً رعایت کنید:
1- هیچ مصراعی نسازید؛ فقط از متن معتبر دیوان حافظ بازیابی کنید.
2- مصراع باید کاملاً مطابق متن اصلی حافظ باشد؛ هیچ واژه یا نشانه‌ای را تغییر ندهید.
<answer>مصراع دوم واقعی حافظ</answer>

مصراع اول:
<first_verse>غرور حسنت اجازت مگر نداد ای گل</first_verse>

نثر بیت:
<couplet_prose>بی‌گمان خودپسندی ناشی از جذابیت ظاهری‌تان مانع شد که جویای وضعیت آن پرنده‌ی دلباخته شوید.</couplet_prose>

**English Translation:**
You are a language model designed to retrieve the exact and authentic second verse of Hafez from his Divan.

You will be given the first verse of a couplet from a Hafez ghazal, followed by a fluent Persian prose rendering of the same couplet. Your task is to identify the true second verse from the Divan of Hafez that corresponds to this couplet, and return it without any modification.

Please strictly follow these instructions:
1- Do not generate any verse; only retrieve it from the authentic text of the Divan of Hafez.
2- The verse must exactly match the original text of Hafez; do not alter any word or punctuation.
<answer>Exact second verse</answer>

Input first verse:
<first__verse>O flower, perhaps the pride of your beauty did not permit it</first__verse>

Couplet prose:
<couplet__prose>Undoubtedly, your pride in your beauty prevented you from asking about the state of that lovesick bird.</couplet__prose>

Table 12: Example prompt for completing the second verse of a ghazal couplet, given the poet's name (Hafez) and a paraphrased version of the couplet's prose.

شما یک مدل زبانی هستید که برای بازیابی مصراع دوم دقیق و اصیل حافظ از دیوان او طراحی شده‌اید.

به شما مصراع اول یکی از ابیات غزل‌های حافظ داده می‌شود و سپس دو واژه از مصراع دوم همان بیت ارائه می‌گردد. وظیفه‌ی شما این است که مصراع دوم واقعی و موجود در دیوان حافظ را که مربوط به همین بیت است، شناسایی کرده و بدون هیچ‌گونه تغییری بازگردانید.

لطفاً دستورالعمل‌های زیر را دقیقاً رعایت کنید:
1- هیچ مصراعی نسازید؛ فقط از متن معتبر دیوان حافظ بازیابی کنید.
2- مصراع باید کاملاً مطابق متن اصلی حافظ باشد؛ هیچ واژه یا نشانه‌ای را تغییر ندهید.
3- خروجی را فقط در قالب زیر بازگردانید و از ارائه هرگونه توضیح اضافی خودداری کنید:
<answer>مصراع دوم واقعی حافظ</answer>

مصراع اول:
<first_verse>غزل گفتی و در سفتی بیا و خوش بخوان حافظ</first_verse>

واژه‌ها:
["نظم", "ثریا"]

**English Translation:**
You are a language model designed to retrieve the exact and authentic second verse of Hafez from his Divan.

You will be given the first verse of a couplet from a Hafez ghazal, followed by two words from the second verse of the same couplet. Your task is to identify the true second verse from the Divan of Hafez that corresponds to this couplet, and return it without any modification.

Please strictly follow these instructions:
1- Do not generate any verse; only retrieve it from the authentic text of the Divan of Hafez.
2- The verse must exactly match the original text of Hafez; do not alter any word or punctuation.
3- Return the output strictly in the following format and do not include any additional text:
<answer>Exact second verse</answer>

Input first verse:
<first__verse>You composed a ghazal and strung its pearls—come, recite it beautifully, O Hafez</first__verse>

Given words:
[”order”, ”Pleiades”]

Table 13: Example prompt for completing the second verse of a ghazal's couplet, given the poet's name (Hafez) and salient words of the second verse.

شما یک مدل زبانی هستید که برای بازیابی مصراع دوم دقیق و اصیل حافظ از دیوان او طراحی شده‌اید.

به شما مصراع اول یکی از ابیات غزل‌های حافظ داده می‌شود و سپس نسخه‌ای از مصراع دوم همان بیت که ترتیب واژگان آن جابه‌جا شده است ارائه می‌گردد. وظیفه‌ی شما این است که مصراع دوم واقعی و موجود در دیوان حافظ را که مربوط به همین بیت است، شناسایی کرده و بدون هیچ‌گونه تغییری بازگردانید.

لطفاً دستورالعمل‌های زیر را دقیقاً رعایت کنید:
1- هیچ مصراعی نسازید؛ فقط از متن معتبر دیوان حافظ بازیابی کنید.
2- مصراع باید کاملاً مطابق متن اصلی حافظ باشد؛ هیچ واژه یا نشانه‌ای را تغییر ندهید.
3- تنها با جابه‌جایی ترتیب واژگان موجود در نسخه‌ی ارائه‌شده از مصراع دوم، آن را به مصراع صحیح برسانید؛ افزودن یا حذف هرگونه واژه یا نشانه به‌هیچ‌وجه مجاز نیست.
4- خروجی را فقط در قالب زیر بازگردانید و از ارائه هرگونه توضیح اضافی خودداری کنید:
<answer>مصراع دوم واقعی حافظ</answer>

نمونه 1:
ورودی:
<first_verse>و گر رسم فنا خواهی که از عالم براندازی</first_verse>
<shuffled_second_verse>فروریزد هر جان برافشان مویت تا ز هزاران </shuffled_second_verse>
خروجی:
<answer>برافشان تا فروریزد هزاران جان ز هر مویت</answer>

نمونه 2:
ورودی:
<first_verse>دو چشم شوخ تو برهم زده خطا و حبش</first_verse>
<shuffled_second_verse>زلف چین هند به و ماچین داده خراج تو </shuffled_second_verse>
خروجی:
<answer>به چین زلف تو ماچین و هند داده خراج</answer>

مصراع اول:
<first_verse>هنگام وداع تو ز بس گریه که کردم</first_verse>

مصراع دوم با ترتیب واژگانی جابه‌جاشده:
<shuffled_second_verse>رخ نور چشم دور نماندست از مرا تو</shuffled_second_verse>

**English Translation:**
You are a language model designed to retrieve the exact and authentic second verse of Hafez from his Divan.

You will be given the first verse of a couplet from a Hafez ghazal, followed by a version of the second verse in which the word order has been shuffled. Your task is to identify the true second verse from the Divan of Hafez that corresponds to this couplet, and return it without any modification.

Please strictly follow these instructions:
1- Do not generate any verse; only retrieve it from the authentic text of the Divan of Hafez.
2- The verse must exactly match the original text of Hafez; do not alter any word or punctuation.
3- Restore the correct verse only by rearranging the order of the words provided in the shuffled version of the second verse; adding or removing any word or punctuation is strictly prohibited.
4- Return the output strictly in the following format and do not include any additional text:
<answer>Exact second verse</answer>

Example 1:
Input:
<first__verse>And if you seek the way of annihilation to cast the world aside</first__verse>
<shuffled__second__verse>collapse every soul scatter your hair from thousands of
</shuffled__second__verse>
Output:
<answer>Scatter your hair so that thousands of souls collapse from each strand</answer>

Example 2:
Input:
<first__verse>Your mischievous eyes have thrown China and Ethiopia into disorder</first__verse>
<shuffled__second__verse>hair China India to and Machin given tribute your
</shuffled__second__verse>
Output:
<answer>To the China of your hair, Machin and India have paid tribute</answer>

Input first verse:
<first__verse>At the moment of parting from you, I wept so intensely</first__verse>

Shuffled second verse:
<shuffled__second__verse>face light eye far remained from me your
</shuffled__second__verse>

Table 14: Example prompt for completing the second verse of a ghazal, given the poet's name (Hafez) and a word-shuffled version of the second verse.

شما یک مدل زبانی هستید که برای بازیابی مصراع دوم دقیق و اصیل حافظ از دیوان او طراحی شده‌اید.

به شما مصراع اول از یکی از ابیات غزل‌های حافظ داده می‌شود. سپس یک پرسش سه‌گزینه‌ای مطرح می‌شود که در آن، مصراع دوم واقعی و موجود در دیوان حافظ، که مربوط به همین مصراع اول است، پرسیده می‌شود.

وظیفه‌ی شما این است که تنها یک گزینه‌ی صحیح را انتخاب کنید. پاسخ شما باید فقط شامل حرف گزینه‌ی درست (الف، ب یا ج) باشد و هیچ متن یا توضیح اضافی ارائه نکنید.

مصراع اول:
<first_verse>خیال زلف تو پختن نه کار هر خامیست</first_verse>

گزینه‌ها:
الف) که زیر سلسله رفتن طریق عیاریست
ب) که این لطیفه عشقم ز رهروی یادست
ج) که رستگاری جاوید در کم آزاریست

**English Translation:**
You are a language model designed to retrieve the exact and authentic second verse of Hafez from his Divan.

You will be given the first verse of a couplet from a Hafez ghazal. You will then be presented with a multiple-choice question asking for the true second verse from the Divan of Hafez that corresponds to this first verse.

Your task is to select only one correct option. Your response must contain only the letter corresponding to the correct option (A, B, or C), and no additional text or explanation.

Input first verse:
<first_verse>To dwell upon your tress is not the work of every novice</first_verse>

Options:
A) For to go beneath the chain is the way of the daring
B) For this subtlety of my love comes from spiritual wayfaring
C) For eternal salvation lies in causing little harm

Table 15: Example prompt for selecting the authentic second verse of a ghazal's couplet, given the poet's name (Hafez)

شما یک مدل زبانی هستید که برای بازیابی مصراع دوم دقیق و اصیل حافظ از دیوان او طراحی شده‌اید.

به شما مصراع اول یکی از ابیات غزل‌های حافظ داده می‌شود و نثر روان فارسی همان بیت نیز ارائه می‌گردد. سپس یک پرسش سه‌گزینه‌ای مطرح می‌شود که در آن، مصراع دوم واقعی و موجود در دیوان حافظ، که مربوط به همین بیت است، پرسیده می‌شود.

وظیفه‌ی شما این است که تنها یک گزینه‌ی صحیح را انتخاب کنید. پاسخ شما باید فقط شامل حرف گزینه‌ی درست (الف، ب یا ج) باشد و هیچ متن یا توضیح اضافی ارائه نکنید.

مصراع اول:
<first_verse>سخن عشق نه آنست که آید به زبان</first_verse>

نثر بیت:
<couplet_prose>سخن عشق نباید که بر زبان آید، پس ای ساقی، می بده و این گفتگو را کوتاه کن.</couplet_prose>

گزینه‌ها:
الف) کاین کرامت سبب حشمت و تمکین من است
ب) ساقیا می ده و کوتاه کن این گفت و شنفت
ج) ناز کم کن که در این باغ بسی چون تو شکفت

**English Translation:**
You are a language model designed to retrieve the exact and authentic second verse of Hafez from his Divan.

You will be given the first verse of a verse from a Hafez ghazal, along with a fluent Persian prose rendering of the same couplet. You will then be presented with a multiple-choice question asking for the true second verse from the Divan of Hafez that corresponds to this couplet.

Your task is to select only one correct option. Your response must contain only the letter corresponding to the correct option (A, B, or C), and no additional text or explanation.

Input first verse:
<first_verse>The discourse of love is not something that can be spoken aloud</first_verse>

Couplet prose:
<couplet_prose>The matter of love should not be spoken openly; therefore, O cupbearer, pour the wine and bring this conversation to an end.</couplet_prose>

Options:
A) For this grace is the source of my dignity and stature
B) O cupbearer, pour the wine and shorten this conversation
C) Be less coquettish, for many like you have blossomed in this garden

Table 16: Example prompt for selecting the authentic second verse of a ghazal's couplet, given the poet's name (Hafez) and the couplet's prose.

شما یک مدل زبانی هستید که برای بازیابی مصراع دوم دقیق و اصیل حافظ از دیوان او طراحی شده‌اید.

به شما مصراع اول یکی از ابیات غزل‌های حافظ داده می‌شود و نثر روان فارسی همان بیت نیز ارائه می‌گردد. سپس یک پرسش سه‌گزینه‌ای مطرح می‌شود که در آن، مصراع دوم واقعی و موجود در دیوان حافظ، که مربوط به همین بیت است، پرسیده می‌شود.

وظیفه‌ی شما این است که تنها یک گزینه‌ی صحیح را انتخاب کنید. پاسخ شما باید فقط شامل حرف گزینه‌ی درست (الف، ب یا ج) باشد و هیچ متن یا توضیح اضافی ارائه نکنید.

مصراع اول:
<first_verse>سخن عشق نه آنست که آید به زبان</first_verse>

نثر بیت:
<couplet_prose>نباید درباره موضوعات عاشقانه صحبتی به میان آید، بنابراین ای جام‌گردان، با فراهم کردن شراب به این بحث خاتمه بده.</couplet_prose>

گزینه‌ها:
الف) کاین کرامت سبب حشمت و تمکین من است
ب) ساقیا می ده و کوتاه کن این گفت و شنفت
ج) ناز کم کن که در این باغ بسی چون تو شکفت

**English Translation:**
You are a language model designed to retrieve the exact and authentic second verse of Hafez from his Divan.

You will be given the first verse of a verse from a Hafez ghazal, along with a fluent Persian prose rendering of the same couplet. You will then be presented with a multiple-choice question asking for the true second verse from the Divan of Hafez that corresponds to this couplet.

Your task is to select only one correct option. Your response must contain only the letter corresponding to the correct option (A, B, or C), and no additional text or explanation.

Input first verse:
<first_verse>The discourse of love is not something that can be spoken aloud</first_verse>

Couplet prose:
<couplet_prose>There should be no discussion of matters of love; therefore, O cupbearer, provide the wine and bring this discussion to an end.</couplet_prose>

Options:
A) For this grace is the source of my dignity and stature
B) O cupbearer, pour the wine and shorten this conversation
C) Be less coquettish, for many like you have blossomed in this garden

Table 17: Example prompt for selecting the authentic second verse of a ghazal's couplet, given the poet's name (Hafez) and a paraphrased version of the couplet's prose.

شما یک مدل زبانی هستید که برای بازیابی مصراع دوم دقیق و اصیل حافظ از دیوان او طراحی شده‌اید.

به شما مصراع اول از یکی از ابیات غزل‌های حافظ داده می‌شود. سپس یک پرسش دوگزینه‌ای مطرح می‌شود که در آن، مصراع دوم واقعی و موجود در دیوان حافظ، که مربوط به همین مصراع اول است، پرسیده می‌شود.

وظیفه‌ی شما این است که تنها یک گزینه‌ی صحیح را انتخاب کنید. پاسخ شما باید فقط شامل حرف گزینه‌ی درست (الف یا ب) باشد و هیچ متن یا توضیح اضافی ارائه نکنید.

مصراع اول:
<first_verse>در گلستان ارم دوش چو از لطف هوا</first_verse>

گزینه‌ها:
الف) زلف سنبل به نسیم سحری می‌آشفت
ب) به زلف نسیم سحری سنبل می‌آشفت

**English Translation:**
You are a language model designed to retrieve the exact and authentic second verse of Hafez from his Divan.

You will be given the first verse of a couplet from a Hafez ghazal. You will then be presented with a two-choice question asking for the true second verse from the Divan of Hafez that corresponds to this first verse.

Your task is to select only one correct option. Your response must contain only the letter corresponding to the correct option (A or B), and no additional text or explanation.

Input first verse:
<first_verse>Last night in the garden of Eram, through the gentleness of the breeze</first_verse>

Options:
A) The hyacinth's tress was tousled by the dawn breeze
B) By the tress of the dawn breeze, the hyacinth was tousled

Table 18: Example prompt for selecting the authentic second verse of a ghazal's couplet, given the poet's name (Hafez) and a word-shuffled second verse as a distractor.

You are a language model designed to retrieve the exact and authentic second line from Shakespeare's Sonnets.

You are given the first line of a couplet from one of Shakespeare's sonnets. Your task is to identify and return the real second line that corresponds to this first line, exactly as it appears in the authoritative text of Shakespeare's Sonnets, without any modification.

Please strictly follow the instructions below:
1- Do not generate or invent any line; retrieve only from the authentic text of Shakespeare's Sonnets.
2- The returned line must exactly match the original wording and punctuation; do not alter, omit, or normalize any word or symbol.
3- Return the output strictly in the following format and do not provide any additional explanation:
<answer>exact second line from Shakespeare's Sonnets</answer>

First line:
<first_line>After my death, dear love, forget me quite,</first_line>

Table 19: Example prompt for completing the second line of a sonnet's couplet, given the poet's name (Shakespeare).

You are a language model designed to retrieve the exact and authentic second line from Shakespeare's Sonnets.

You are given the first line of a couplet from one of Shakespeare's sonnets, followed by two words taken from the second line of the same couplet. Your task is to identify and return the real second line that corresponds to this couplet, exactly as it appears in the authoritative text of Shakespeare's Sonnets, without any modification.

Please strictly follow the instructions below:
1- Do not generate or invent any line; retrieve only from the authentic text of Shakespeare's Sonnets.
2- The returned line must exactly match the original wording and punctuation; do not alter, omit, or normalize any word or symbol.
3- Return the output strictly in the following format and do not provide any additional explanation:
<answer>verbatim second line from Shakespeare's Sonnets</answer>

First line:
<first_line>Your monument shall be my gentle verse,</first_line>

Words:
[”eyes”, ”o’er-read”]

Table 20: Example prompt for completing the second line of a sonnet's couplet, given the poet's name (Shakespeare) and salient words of the second line.

You are a language model designed to retrieve the exact and authentic second line from Shakespeare's Sonnets.

You are given the first line of a couplet from one of Shakespeare's sonnets, followed by a version of the second line from the same couplet in which the word order has been shuffled. Your task is to identify and return the real second line that corresponds to this couplet, exactly as it appears in the authoritative text of Shakespeare's Sonnets, without any modification.

Please strictly follow the instructions below:
1- Do not generate or invent any line; retrieve only from the authentic text of Shakespeare's Sonnets.
2- The returned line must be completely identical to the original line in wording and punctuation; do not change, normalize, add, or remove any word or symbol.
3- Restore the correct second line solely by reordering the words present in the provided shuffled version; adding or deleting any word or symbol is strictly prohibited.
4- Return the output strictly in the following format and do not provide any additional explanation:
<answer>exact second line from Shakespeare's Sonnets</answer>

Example 1:
Input:
<first_line>Rough winds do shake the darling buds of May,</first_line>
<shuffled_second_line>lease summer's short and too all a date: hath</shuffled_second_line>
Output:
<answer>And summer's lease hath all too short a date:</answer>

Example 2:
Input:
<first_line>Love alters not with his brief hours and weeks,</first_line>
<shuffled_second_line>it to of bears but edge even the doom. out</shuffled_second_line>
Output:
<answer>But bears it out even to the edge of doom.</answer>

First line:
<first_line>As victors of my silence cannot boast;</first_line>

Second line with shuffled word order:
<shuffled_second_line>not was fear I any of from thence: sick</shuffled_second_line>

Table 21: Example prompt for completing the second line of a sonnet's couplet, given the poet's name (Shakespeare) and a word-shuffled version of the second line.

# D Failure Modes

**Verse completion.** Table 22 provides examples of failure categories observed in our LLM-generated data for completion experiments. Hallucination is when LLMs access a plausible but incorrect verse that does not exist. Prosaic Paraphrase is when models access an explanation of the verse's meaning in prose format. Incomplete Completion is when models access only a part of the correct verse. Contextual Confusion is when models access a verse from another ghazal of the same poet.

**Verse recognition.** Figures 2 and 3 present the distribution of distractor options for the recognition-based task across different scenarios in the Hafez and Saadi corpora, respectively.

# E Corpus Statistics: Persian vs. English

Table 23 compares the Persian ghazal couplets and English sonnet couplets used in our cross-lingual analysis. All Persian text was normalized using the pipeline described in Section 4.3; English text was lowercased and stripped of punctuation before tokenisation. Character lengths and word counts are computed per verse/line, then averaged across the corpus.

# F Pairwise McNemar Test Results

We perform pairwise McNemar tests (McNemar, 1947) between each pair of models for the completion and recognition tasks to assess whether the observed differences are statistically significant. Tables 24 and 25 summarize the resulting comparisons. Most pairwise differences across the full set of models are statistically significant, although smaller differences are not consistently significant. For example, the two Gemma 3 variants show largely nonsignificant differences, whereas GPT-5.2 differs significantly from the other models across nearly all conditions. We examine the comparison between DeepSeek V3.2 and DeepSeek R1 in greater detail below.

**DeepSeek V3.2 vs. DeepSeek R1.** For completion, the differences between the two models are statistically significant in only a subset of settings. For Hafez, significant differences are observed under the baseline, prose-mediated, and salient-word settings ($p < 0.05$), while the paraphrase-based setting does not yield a significant difference. For

| Failure Category | Model | Scenario | Expected Output | Model Output |
|---|---|---|---|---|
| Hallucination | GPT-5.2 | Baseline | سهی قدان سیه چشم ماه سیما را | چرا ز دوست به دشمن چنین جدایی نیست |
| Hallucination | Gemma 3 12B | +Prose | تفقـدی نکند طوطی شکرخا را | چون تو نیایی دلبری چه سود مرا |
| Prosaic Paraphrase | Claude Sonnet 4.5 | +Prose | به یاد دار محبان باده پیما را | ز عاشقان بی‌بهره خود یادی کن |
| Prosaic Paraphrase | DeepSeek V3.2 | +Para. | نمی‌کنی به ترحم، نِطاق سلسله سست | تو همچنان بی‌شفقی و از جور نمی‌کاهی |
| Incomplete Completion | DeepSeek R1 | +Salient | زلف سنبل به نسیم سحری می‌آشفت | زلف سنبل به شکن بر شاخ گل می‌زد خوشه |
| Incomplete Completion | Gemma 3 27B | +Shuffle | که کس نگشود و نگشاید به حکمت این معما را | و کس نگشاید حکمت این معما را |
| Contextual Confusion | Llama 4 Scout | Name | تبارک الله از این فتنه‌ها که در سر ماست | که در طریق عشقبازی امن و آسایش بلاست |
| Contextual Confusion | DeepSeek R1 | Name | چو حافظ بنده و هندوی فرخ | ز هر چه رنگ تعلق پذیرد آزاد است |

Table 22: Illustrative failure modes observed across models and completion scenarios.

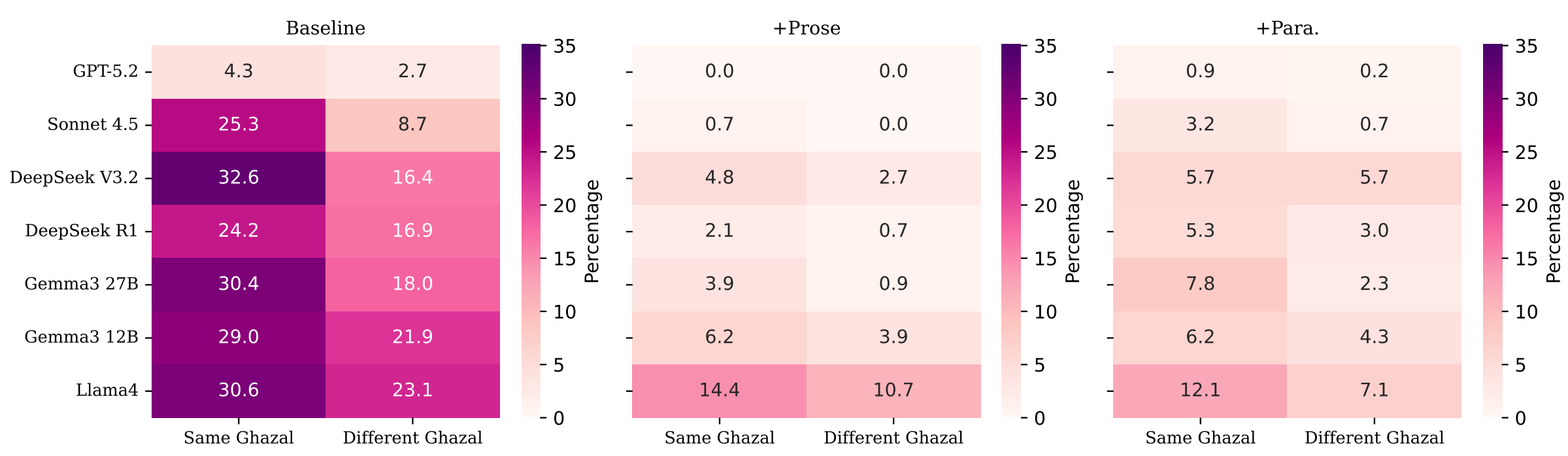

Figure 2: Percentage distribution of model selections among distractor options in the multiple-choice verse recognition task under varying contextual cues for Hafez's ghazals.

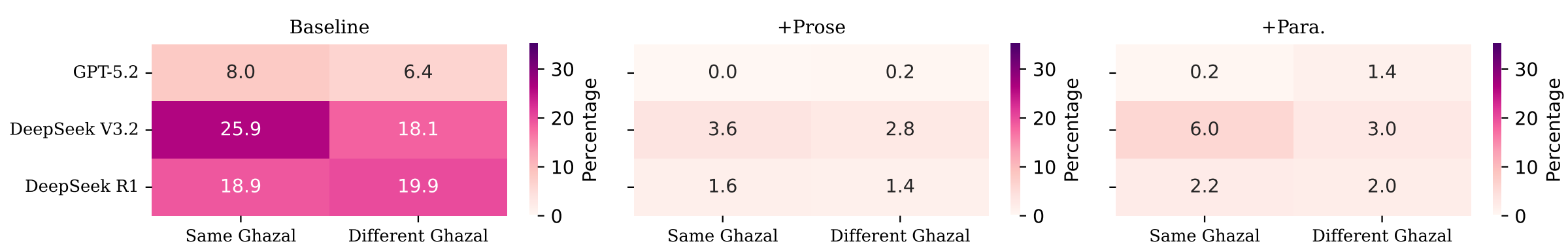

Figure 3: Percentage distribution of model selections among distractor options in the multiple-choice verse recognition task under varying contextual cues for Saadi's ghazals.

| Statistic | Persian Ghazals | English Sonnets |
|---|---|---|
| Number of Couplets | 936 | 350 |
| **Length (characters)** | | |
| First Verse/Line (mean ± std) | 30.6 ± 4.5 | 42.4 ± 3.8 |
| Second Verse/Line (mean ± std) | 31.1 ± 4.5 | 42.3 ± 4.2 |
| **Word Count** | | |
| First Verse/Line (mean ± std) | 7.4 ± 1.5 | 8.3 ± 1.1 |
| Second Verse/Line (mean ± std) | 7.5 ± 1.4 | 8.1 ± 1.1 |
| **Lexical Diversity** | | |
| Vocabulary Size (unique words) | 3,344 | 1,696 |
| Type–Token Ratio (TTR) | 0.24 | 0.30 |
| Hapax Legomena Ratio | 0.14 | 0.19 |

Table 23: Structural statistics for the Persian ghazals and English sonnets.

Saadi, the two models exhibit largely similar success patterns across completion scenarios, with no significant differences observed. In the Saadi baseline setting, both models produce identical completion outcomes for all instances, making McNemar's test undefined.

Recognition shows a stronger distinction between the two models. For Hafez, statistically significant differences are observed under the baseline, prose-mediated, and shuffled settings, while the paraphrase-based setting does not reach significance ($p = 0.066$). For Saadi, all recognition scenarios exhibit significant differences, with the strongest differences observed under paraphrase-based and shuffled settings ($p < 0.001$). These results indicate that the two models' success patterns differ more consistently for recognition than for completion.

| Comparison | Completion | | | | |
|---|---|---|---|---|---|
| | **Baseline** | **+Prose** | **+Para.** | **+Salient** | **+Shuffle** |
| **Hafez** | | | | | |
| GPT-5.2 vs. Sonnet 4.5 | 177.0 ($p < .001^{***}$) | 185.4 ($p < .001^{***}$) | 169.1 ($p < .001^{***}$) | 122.2 ($p < .001^{***}$) | 22.8 ($p < .001^{***}$) |
| GPT-5.2 vs. DeepSeek V3.2 | 158.1 ($p < .001^{***}$) | 186.2 ($p < .001^{***}$) | 166.1 ($p < .001^{***}$) | 130.1 ($p < .001^{***}$) | 25.0 ($p < .001^{***}$) |
| GPT-5.2 vs. DeepSeek R1 | 145.3 ($p < .001^{***}$) | 168.6 ($p < .001^{***}$) | 159.1 ($p < .001^{***}$) | 101.0 ($p < .001^{***}$) | 19.3 ($p < .001^{***}$) |
| GPT-5.2 vs. Gemma 3 27B | 192.0 ($p < .001^{***}$) | 223.0 ($p < .001^{***}$) | 192.0 ($p < .001^{***}$) | 145.0 ($p < .001^{***}$) | 89.3 ($p < .001^{***}$) |
| GPT-5.2 vs. Gemma 3 12B | 192.0 ($p < .001^{***}$) | 223.0 ($p < .001^{***}$) | 193.0 ($p < .001^{***}$) | 146.0 ($p < .001^{***}$) | 98.0 ($p < .001^{***}$) |
| GPT-5.2 vs. Llama 4 | 190.0 ($p < .001^{***}$) | 217.0 ($p < .001^{***}$) | 187.0 ($p < .001^{***}$) | 141.1 ($p < .001^{***}$) | 86.5 ($p < .001^{***}$) |
| Sonnet 4.5 vs. DeepSeek v3.2 | 6.3 ($.012^{*}$) | .0 (1.0) | .1 (.700) | 1.0 (.327) | .0 (1.0) |
| Sonnet 4.5 vs. DeepSeek R1 | 17.4 ($p < .001^{***}$) | 5.3 ($.021^{*}$) | 3.6 (.059) | 6.6 ($.010^{*}$) | .2 (.635) |
| Sonnet 4.5 vs. Gemma 3 27B | 11.5 ($p < .001^{***}$) | 28.0 ($p < .001^{***}$) | 17.1 ($p < .001^{***}$) | 15.4 ($p < .001^{***}$) | 38.9 ($p < .001^{***}$) |
| Sonnet 4.5 vs. Gemma 3 12B | 11.5 ($p < .001^{***}$) | 28.0 ($p < .001^{***}$) | 18.0 ($p < .001^{***}$) | 18.0 ($p < .001^{***}$) | 49.0 ($p < .001^{***}$) |
| Sonnet 4.5 vs. Llama 4 | 9.6 ($.002^{**}$) | 16.5 ($p < .001^{***}$) | 7.0 ($.007^{**}$) | 13.5 ($p < .001^{***}$) | 34.0 ($p < .001^{***}$) |
| DeepSeek V3.2 vs. DeepSeek R1 | 4.8 ($.029^{*}$) | 6.5 ($.011^{*}$) | 2.6 (.110) | 15.7 ($p < .001^{***}$) | 0.5 (.496) |
| DeepSeek V3.2 vs. Gemma 3 27B | 26.3 ($p < .001^{***}$) | 27.3 ($p < .001^{***}$) | 18.4 ($p < .001^{***}$) | 9.6 ($.002^{**}$) | 36.5 ($p < .001^{***}$) |
| DeepSeek V3.2 vs. Gemma 3 12B | 26.3 ($p < .001^{***}$) | 27.3 ($p < .001^{***}$) | 21.0 ($p < .001^{***}$) | 12.1 ($p < .001^{***}$) | 48.0 ($p < .001^{***}$) |
| DeepSeek V3.2 vs. Llama 4 | 22.8 ($p < .001^{***}$) | 18.6 ($p < .001^{***}$) | 12.2 ($p < .001^{***}$) | 6.7 ($.009^{**}$) | 34.2 ($p < .001^{***}$) |
| DeepSeek R1 vs. Gemma 3 27B | 39.0 ($p < .001^{***}$) | 43.0 ($p < .001^{***}$) | 27.3 ($p < .001^{***}$) | 33.0 ($p < .001^{***}$) | 43.9 ($p < .001^{***}$) |
| DeepSeek R1 vs. Gemma 3 12B | 39.0 ($p < .001^{***}$) | 41.2 ($p < .001^{***}$) | 30.0 ($p < .001^{***}$) | 34.0 ($p < .001^{***}$) | 54.0 ($p < .001^{***}$) |
| DeepSeek R1 vs. Llama 4 | 33.6 ($p < .001^{***}$) | 32.1 ($p < .001^{***}$) | 19.5 ($p < .001^{***}$) | 29.3 ($p < .001^{***}$) | 41.4 ($p < .001^{***}$) |
| Gemma 3 27B vs. Gemma 3 12B | – (N/A) | .5 (.479) | .0 (1.0) | .0 (1.0) | 2.3 (.131) |
| Gemma 3 27B vs. Llama 4 | .2 (.617) | 3.1 (.077) | 2.3 (.131) | .2 (.617) | .0 (1.0) |
| Gemma 3 12B vs. Llama 4 | .2 (.617) | 4.2 ($.041^{*}$) | 4.2 ($.041^{*}$) | 1.3 (.248) | 4.2 ($.041^{*}$) |
| **Saadi** | | | | | |
| GPT-5.2 vs. DeepSeek V3.2 | 9.1 ($.003^{*}$) | 47.2 ($p < .001^{***}$) | 31.0 ($p < .001^{***}$) | 33.0 ($p < .001^{***}$) | 31.6 ($p < .001^{***}$) |
| GPT-5.2 vs. DeepSeek R1 | 9.1 ($.003^{*}$) | 43.2 ($p < .001^{***}$) | 28.0 ($p < .001^{***}$) | 34.0 ($p < .001^{***}$) | 14.9 ($p < .001^{***}$) |
| DeepSeek V3.2 vs. DeepSeek R1 | – (N/A) | 1.1 (.289) | 1.3 (.248) | .0 (1.0) | 2.5 (.112) |

Table 24: Pairwise McNemar tests comparing each pair of models across completion scenarios. Each cell reports the McNemar $\chi^2$ statistic followed by the corresponding two-sided $p$-value in parentheses. Significance levels: * $p < 0.05$, ** $p < 0.01$, *** $p < 0.001$.

| Comparison | Recognition | | | |
| --- | --- | --- | --- | --- |
| | **Baseline** | **+Prose** | **+Para.** | **+Shuffle** |
| **Hafez** | | | | |
| GPT-5.2 vs. Sonnet 4.5 | 97.8 ($p < .001^{***}$) | 1.3 ($.248$) | 8.6 ($.003^{**}$) | 13.1 ($p < .001^{***}$) |
| GPT-5.2 vs. DeepSeek V3.2 | 169.1 ($p < .001^{***}$) | 31.0 ($p < .001^{***}$) | 38.0 ($p < .001^{***}$) | 84.3 ($p < .001^{***}$) |
| GPT-5.2 vs. DeepSeek R1 | 134.4 ($p < .001^{***}$) | 10.1 ($.001^{**}$) | 27.3 ($p < .001^{***}$) | 29.3 ($p < .001^{***}$) |
| GPT-5.2 vs. Gemma 3 27B | 166.2 ($p < .001^{***}$) | 19.0 ($p < .001^{***}$) | 33.6 ($p < .001^{***}$) | 52.2 ($p < .001^{***}$) |
| GPT-5.2 vs. Gemma 3 12B | 170.5 ($p < .001^{***}$) | 42.0 ($p < .001^{***}$) | 32.7 ($p < .001^{***}$) | 148.0 ($p < .001^{***}$) |
| GPT-5.2 vs. Llama 4 | 180.7 ($p < .001^{***}$) | 108.0 ($p < .001^{***}$) | 73.3 ($p < .001^{***}$) | 91.2 ($p < .001^{***}$) |
| Sonnet 4.5 vs. DeepSeek v3.2 | 37.1 ($p < .001^{***}$) | 24.7 ($p < .001^{***}$) | 21.8 ($p < .001^{***}$) | 53.6 ($p < .001^{***}$) |
| Sonnet 4.5 vs. DeepSeek R1 | 6.2 ($.013^{*}$) | 4.9 ($.026^{*}$) | 9.8 ($.002^{**}$) | 3.8 ($.049^{*}$) |
| Sonnet 4.5 vs. Gemma 3 27B | 20.3 ($p < .001^{***}$) | 13.1 ($p < .001^{***}$) | 17.3 ($p < .001^{***}$) | 19.6 ($p < .001^{***}$) |
| Sonnet 4.5 vs. Gemma 3 12B | 20.0 ($p < .001^{***}$) | 35.6 ($p < .001^{***}$) | 15.4 ($p < .001^{***}$) | 102.1 ($p < .001^{***}$) |
| Sonnet 4.5 vs. Llama 4 | 33.8 ($p < .001^{***}$) | 101.2 ($p < .001^{***}$) | 55.1 ($p < .001^{***}$) | 56.9 ($p < .001^{***}$) |
| DeepSeek V3.2 vs. DeepSeek R1 | 8.1 ($.004^{**}$) | 10.3 ($.001^{**}$) | 3.4 ($.066$) | 35.2 ($p < .001^{***}$) |
| DeepSeek V3.2 vs. Gemma 3 27B | .0 ($.882$) | 2.6 ($.105$) | .3 ($.556$) | 9.4 ($.002^{**}$) |
| DeepSeek V3.2 vs. Gemma 3 12B | .2 ($.654$) | 1.5 ($.222$) | .1 ($.746$) | 16.9 ($p < .001^{***}$) |
| DeepSeek V3.2 vs. Llama 4 | 1.7 ($.196$) | 58.3 ($p < .001^{***}$) | 13.3 ($p < .001^{***}$) | .2 ($.618$) |
| DeepSeek R1 vs. Gemma 3 27B | 5.1 ($.024^{*}$) | 2.6 ($.110$) | .8 ($.358$) | 6.6 ($.010^{*}$) |
| DeepSeek R1 vs. Gemma 3 12B | 7.8 ($.005^{**}$) | 20.9 ($p < .001^{***}$) | 1.2 ($.282$) | 77.8 ($p < .001^{***}$) |
| DeepSeek R1 vs. Llama 4 | 14.1 ($p < .001^{***}$) | 85.5 ($p < .001^{***}$) | 24.5 ($p < .001^{***}$) | 33.6 ($p < .001^{***}$) |
| Gemma 3 27B vs. Gemma 3 12B | .6 ($.455$) | 10.8 ($.001^{**}$) | .0 ($.897$) | 59.2 ($p < .001^{***}$) |
| Gemma 3 27B vs. Llama 4 | 2.7 ($.098$) | 69.8 ($p < .001^{***}$) | 17.7 ($p < .001^{***}$) | 14.7 ($p < .001^{***}$) |
| Gemma 3 12B vs. Llama 4 | .6 ($.432$) | 35.8 ($p < .001^{***}$) | 14.9 ($p < .001^{***}$) | 16.8 ($p < .001^{***}$) |
| **Saadi** | | | | |
| GPT-5.2 vs. DeepSeek V3.2 | 112.8 ($p < .001^{***}$) | 27.3 ($p < .001^{***}$) | 27.6 ($p < .001^{***}$) | 167.4 ($p < .001^{***}$) |
| GPT-5.2 vs. DeepSeek R1 | 83.2 ($p < .001^{***}$) | 10.6 ($.001^{**}$) | 6.3 ($.012^{*}$) | 33.1 ($p < .001^{***}$) |
| DeepSeek V3.2 vs. DeepSeek R1 | 4.3 ($.039^{*}$) | 8.3 ($.004^{**}$) | 12.0 ($p < .001^{***}$) | 109.4 ($p < .001^{***}$) |

Table 25: Pairwise McNemar tests comparing each pair of models across recognition scenarios. Each cell reports the McNemar $\chi^2$ statistic followed by the corresponding two-sided $p$-value in parentheses. Significance levels: * $p < 0.05$, ** $p < 0.01$, *** $p < 0.001$.